\documentclass{article}
\PassOptionsToPackage{numbers,compress}{natbib}
\usepackage[main,final]{neurips_2026}
\usepackage[utf8]{inputenc}
\usepackage[T1]{fontenc}
\usepackage{amsmath,amssymb,amsfonts}
\usepackage{graphicx,booktabs,makecell,multirow}
\usepackage{microtype,xcolor,caption,url}

\usepackage{hyperref}
\usepackage{algorithm,algpseudocode}
\usepackage[nameinlink,noabbrev]{cleveref}

\hypersetup{hidelinks,pdftitle={Sparse Sum-of-Squares Layers for Certified Learning},
pdfauthor={Ege Gursoy and Nicolas Mansard}}

\title{Sparse Sum-of-Squares Layers\\for Certified Learning}

\author{
	Ege Gursoy$^1$ \quad 
	Nicolas Mansard$^{1,2}$ \\ \\
	{$^1$ LAAS-CNRS} \quad
	{$^2$ ANITI}
}

\begin{document}
\maketitle
\begin{abstract}
A central limitation of modern machine learning systems is the lack of formal certificates guaranteeing their behavior beyond empirical, sample-based evaluation. Sum-of-squares (SOS) methods provide such guarantees by certifying polynomial inequalities over continuous domains via semidefinite programming (SDP). While prior work has introduced SDP layers as differentiable components in neural networks, repeated learning calls can lose the computational benefit of the forward program's sparse representation. We propose \textit{Sparse SOS Layers}, differentiable value layers whose backward passes reuse the local coordinates of the forward solve. Under a fixed template and learned objective coefficients, each layer avoids building dense certificates and reuses its sparse structure across training steps. Compared with existing methods, our approach achieves up to $2.7\times$ speedup in total warm-call time, with nearly $100\times$ lower memory usage ($1.7$ vs. $160$ GiB) for relatively large problem sizes for the tested dense parameterization. Experiments cover sparse polynomial certification, pendulum control, combinatorial relaxations, robotics projection, oscillator chains, and pose graph synchronization. A design study achieves zero certified violations across unseen problem instances and sizes, while requiring roughly $7\times$ fewer optimization solves than estimating gradients by perturbing the objective coefficients and re-solving the problem (\textit{finite differences}). These results show that sparse certificate values can be effectively integrated into learning pipelines when the backward pass respects the verifier's structure.
\end{abstract}
\section{Introduction}
\label{sec:intro}
Many learning systems are trained from finite data but used under continuous constraints that pointwise losses are hard to enforce. Certificates address this by converting such claims into checkable objects. Sum-of-squares (SOS) certificates provide one standard approach: a polynomial written as a sum of squares is guaranteed to be nonnegative. With a fixed monomial basis, searching for this representation becomes a semidefinite program (SDP) over positive semidefinite (PSD) matrices \citep{parrilo2003sdp,lasserre2001moments,lasserre2008gpm,lasserre2010momentsbook,laurent2009sos,lasserre2015introduction,nie2023moment,theobald2024real}. This makes SOS useful for polynomial constraints in control \citep{papachristodoulou2005sostools,prajna2004safety,henrion2020momentsos}, and also connects to SDP relaxations for quadratic and low-order polynomial optimization problems \citep{shor1987quadratic,luo2010semidefinite}.

The difficulty is that dense SOS programs grow quickly with dimension\footnote{Dense SOS uses a PSD matrix of size $\binom{n+r}{r}$ for degree $2r$ certificates in $n$ variables.}.
The key to scaling SOS to relevant size is to exhibit and exploit sparsity in the polynomial structure \citep{waki2006sparse,magron2023sparse,wang2022cstssos}. A dense matrix treats all variables as coupled, even when the underlying problem is local. Sparse SOS methods avoid this by exploiting structure that appears in chains, grids, robots, and graphs. Correlative sparsity limits each polynomial term to a small set of variables, while chordal sparsity replaces one large PSD matrix by smaller overlapping blocks tied by consistency constraints \citep{grone1984pdcompletion,fukuda2001matrix1,nakata2003matrix2,waki2006sparse,vandenberghe2015chordal,zheng2020chordal}. This sparse representation is what makes the forward certificate solve practical in applications such as vision, perception, and robotics \citep{rosen2019se,tian2021distributed,ahmadi2016geometry,holmes2025sdprlayers}.

An appealing research direction in ML is to leverage positivity certificates by treating an SOS formulation as a differentiable optimization layer inside a neural network. Then one can learn by optimizing through its output, following the broader line of differentiable optimization layers \citep{amos2017optnet,agrawal2019dco,agrawal2019cone,blondel2022implicit,besancon2024diffopt,gould2022ddn}. A generic differentiable optimization layer differentiates the canonical cone program, which may lose the local block structure that made the forward solve tractable \citep{agrawal2019dco}. Prior work trains reinforcement-learning controllers using gradients from SOS-based certificate optimization~\citep{wang2023joint}. Conic differentiation has also been used to optimize spatio-temporal decompositions of control-related SOS programs~\citep{cibulka2024decomposition}.

This paper asks whether that same sparse certificate can also be used during training. We consider a fixed sparse certificate template whose coefficients are predicted by a neural network. The layer solves the resulting sparse SDP and returns a scalar certified value, such as a margin. The training loss then differentiates through this solve. This is different from two common alternatives, which are: \textit{sampled surrogate} which checks the certificate inequality only on finitely many points, and \textit{stop gradient} method which solves the certificate forward but removes its gradient from learning. Both can be useful, but neither trains directly through the deployed sparse certificate.

Our contribution is to make the backward pass use the same sparse blocks as the forward certificate solve. The result is a sparse differentiable layer for structured SDP certificates, covering both quadratic relaxations and SOS certificates used for local polynomial constraints. For sparse SOS, the backward pass must preserve the representation that made the forward solve tractable. SDPRLayers \citep{holmes2025sdprlayers} studies certifiable backpropagation through polynomial optimization problems in robotics using Shor relaxations, where products of variables are lifted into a PSD matrix and the rank constraint is dropped. CVXPYLayers provides a general differentiable convex programming interface and can also represent explicitly chordal models. Our layer targets the sparse certificate representation directly.

When the loss depends only on a scalar certified value, the gradient follows an envelope rule and contracts the optimal sparse blocks with the derivatives of the predicted coefficients. No explicit dense certificate needs to be reconstructed. These are standard ideas from sensitivity analysis, perturbation analysis, and value-function differentiation \citep{fiacco1983sensitivity,bonnans2000perturbation,mehmood2020valuefunction}. Our contribution is the fixed-template architecture that discovers the clique structure and objective feature map once, keeps the solver's local coordinates across the automatic differentiation boundary, and updates only the parameter-dependent objective coefficients between calls.

Our work is related to several approaches that reduce, replace, or complement exact SOS, including DSOS and SDSOS inner approximations \citep{ahmadi2019dsos}, basis pursuit and column generation \citep{ahmadi2016basispursuit,ahmadi2017colgen}, decomposed structured cones \citep{miller2022decomposed,zheng2018ssosbridge}, sample-based variants \citep{lofberg2004coefficients,cifuentes2017sampling}, and kernel or infinite-dimensional alternatives \citep{rudi2025kernelmin,berthier2022infinite,groudiev2025sampling}. Neural Lyapunov methods also use Lyapunov conditions as learning or verification signals \citep{chang2019neurallyapunov,zhou2022unknown,wu2023discrete,yang2024lyapunovstable,shi2024ctbab,korda2022verification}. These lines address different parts of the certificate problem. We focus on the case where the deployed verifier is already a sparse SOS program, and ask whether the solved relaxation can remain in the training loop without paying for dense backward algebra.

Our contributions are,
    (i) we define a differentiable sparse SOS value layer for fixed certificate templates and learned objective coefficients under standard regularity assumptions;
    (ii) we give a count-based cost statement for preserving local solver coordinates across repeated differentiation, supported by controlled sparse-baseline and large-scale comparisons;
    (iii) we evaluate the layer on sparse polynomial and semidefinite problems, pendulum control, and a design task that tests whether optimizing the certified value changes the learned outcome.
\section{Method}
\label{sec:method}

A polynomial certificate proves that a polynomial inequality holds on a region. SOS gives a tractable sufficient condition for such a proof: if $p(\xi)$ can be written as $\sum_i q_i(\xi)^2$, then $p(\xi)\ge 0$. With a fixed monomial basis $m(\xi)$, finding the squared polynomials $q_i$ is equivalent to finding a PSD matrix $Q$ such that $p(\xi)=m(\xi)^\top Qm(\xi)$. $Q$ is often called a Gram matrix~\citep{parrilo2003sdp,lasserre2001moments} and matching polynomial coefficients gives linear equations in $Q$. Sparse SOS formulations use local Gram matrices over small variable sets. Their dual moment formulations use local moment blocks with equality constraints on shared coordinates. \textit{Sparse SOS Layers} uses an objective-parametrized conic form to differentiate the optimal value while retaining these local coordinates.

\subsection{Problem definition}
\label{subsec:problem_setting}

Let $x$ denote an input, and let a neural model with parameters $\theta$ produce the coefficients of a sparse polynomial certificate problem. After fixing a relaxation order, monomial basis, and sparse decomposition, the forward certificate layer solves a sparse conic program of the form
\begin{equation}
\label{eq:sparse_sos_primal}
v_\theta(x)
~=~
\min_{y = (y_c)_{c \in \mathcal C}}
\sum_{c \in \mathcal C} \langle c_c(\theta,x), y_c \rangle
\quad
\text{s.t.}
\quad
Fy = d,
~
y_c \in \mathcal K_c \;\; \forall c \in \mathcal C
\end{equation}
where $\mathcal C$ indexes the maximal cliques of the sparse SOS decomposition. A clique is a local group of variables that appear together in the same part of the polynomial, such as neighboring nodes in a chain or adjacent links in a robot. The vector $y_c$ collects the local conic block coordinates associated with clique $c$, and $\mathcal K_c$ denotes the corresponding cone. For a semidefinite block, membership in $\mathcal K_c$ means that the matrix represented by these coordinates is PSD. The constraint $Fy=d$ collects the fixed affine equations, including moment identities and separator consistency, while $c_c(\theta,x)$ denotes the objective coefficients exposed by the upstream model. The learned coefficients enter through $c_c(\theta,x)$ while the feasible set is fixed locally. Moment and Shor relaxations naturally have this form as SOS Gram formulations with coefficient matching in the right-hand side are used through the equivalent dual value form for this pullback.

The key point is that \eqref{eq:sparse_sos_primal} is \emph{not} treated as a dense ambient SDP, meaning we do not assemble the single large SDP that would contain all variables and all monomials at once. The decomposition into clique-local blocks and overlap constraints is the representation used throughout the layer. Thus, the certified scalar may be a margin, lower bound, or certificate residual used directly in the loss, with the local optimal blocks retained for the value pullback.

\subsection{Local blocks and shared moments}
\label{subsec:local_blocks}
The polynomial's interaction pattern determines the local blocks and their overlaps. Consider a polynomial in $n\geq3$ variables with chain-structured interactions
\begin{equation}
p_\theta(\xi)=1+\sum_{i=1}^n\bigl(\xi_i^4+a_i(\theta)\xi_i\bigr)
+\sum_{i=1}^{n-1}b_{i,i+1}(\theta)\xi_i\xi_{i+1}
\label{eq:polynomial_chain}
\end{equation}
where $\xi$ denotes the polynomial variables and $a_i(\theta)$ and $b_{i,i+1}(\theta)$ are learned coefficients. The fixed interaction pattern has maximal cliques $C_i=\{\xi_i,\xi_{i+1}\}$ for $i=1,\ldots,n-1$. Adjacent cliques share the separator $C_i\cap C_{i+1}=\{\xi_{i+1}\}$. Each clique uses the order-two basis $(1,\xi_i,\xi_{i+1},\xi_i^2,\xi_i\xi_{i+1},\xi_{i+1}^2)$, giving $n-1$ local moment blocks of size $6\times6$.

A moment $L(\xi^\alpha)$ is the coordinate assigned to monomial $\xi^\alpha$. Each block collects the moments of products of its basis monomials. Moments involving only a shared variable appear in both adjacent blocks; the separator constraints in $Fy=d$ require these copies to agree. The bases, local blocks, and fixed affine constraints form the certificate template. Updating $a_i(\theta)$ and $b_{i,i+1}(\theta)$ changes the objective coefficients in \eqref{eq:sparse_sos_primal} while preserving this template. The scalar derivative below uses the local coordinates returned by that solve.

\subsection{Scalar value differentiation}
\label{subsec:value_only}

Generic differentiable optimization layers provide derivatives of the optimal solution with respect to problem data~\citep{agrawal2019dco,agrawal2019cone,holmes2025sdprlayers}. Such solution-map derivatives support losses that depend on the optimizer itself. The backward computation can apply the required Jacobian action without explicitly forming a full Jacobian, and the forward program can use a chordal formulation. Forward sparsity and the choice of derivative computation are therefore separate questions.

Our loss depends on the certified scalar, $\mathcal L(\theta;x)=\phi(v_\theta(x))$. This permits a value-function sensitivity rule that uses the returned optimizer directly~\citep{fiacco1983sensitivity,bonnans2000perturbation,mehmood2020valuefunction}. Let $y^\star(\theta,x)$ solve \eqref{eq:sparse_sos_primal}, and define the coefficient Jacobian $J_{c_c}(\theta,x)=\partial c_c(\theta,x)/\partial\theta$. Under the fixed-feasible-set and regularity assumptions in \Cref{app:theory}, a locally unique optimizer gives
\begin{equation}
\nabla_\theta v_\theta(x)
=
\sum_{c\in\mathcal C}
J_{c_c}(\theta,x)^\top y_c^\star(\theta,x)
\label{eq:value_gradient}
\end{equation}
The envelope rule accounts for the effect of optimizing over $y$: the first-order change in the optimal value is obtained by changing the objective coefficients at the current optimum. Consequently, \eqref{eq:value_gradient} requires the optimal coordinates $y_c^\star$, without differentiating their motion with $\theta$. For the chain in \Cref{subsec:local_blocks}, the derivative with respect to $b_{i,i+1}$ is the returned moment $L^\star(\xi_i\xi_{i+1})$

For an incoming scalar gradient $g=\partial\mathcal L/\partial v$, the backward rule is
\begin{equation}
\label{eq:value_pullback}
\nabla_\theta \mathcal L
=
g \sum_{c\in\mathcal C}
J_{c_c}(\theta,x)^\top y_c^\star
\end{equation}
This reverse-mode operation, also called a \emph{pullback}, maps the gradient at the layer output to a gradient with respect to its parameters. Holding the returned coordinates fixed while evaluating \eqref{eq:value_pullback} implements the envelope derivative. The interface ensures the loss is evaluated at the optimal value, leaving the gradient formula to define the final sensitivity rule. When the optimizer is not unique, we interpret the value map through its subgradient set and use the optimal coordinates returned by the solver. 

Throughout this work, \textit{exact} refers to computing the derivative of the solved conic relaxation up to solver tolerance


\subsection{Preserving sparse coordinates across the layer}
\label{subsec:fixed_template}
Sparse SOS engines already discover local bases and clique structure and assemble the corresponding conic program~\citep{waki2006sparse,wang2021tssos}. The envelope rule is also established. \textit{Sparse SOS Layers} adds a fixed-template interface that preserves the association between polynomial objective coefficients and solver coordinates through repeated forward and backward calls. This association lets the layer update the sparse model and evaluate its value derivative using the same stored feature map.

Let $\eta_\theta(x)$ be the vector of exposed objective coefficients. Write their placement in each local block as
\begin{equation}
c_c(\theta,x)=c_c^0+B_c\eta_\theta(x)
\label{eq:coefficient_map}
\end{equation}
where $c_c^0$ contains fixed objective terms and $B_c$ describes the fixed sparse coefficient map. The entries of $B_c$ identify which objective coordinates each coefficient affects, including their weights. This notation describes the stored feature map; it does not require a dense matrix.

At setup, the layer retains these coefficient-to-coordinate associations together with the monomial bases, clique layout, and overlap equations. A forward call updates the affected objective entries, solves \eqref{eq:sparse_sos_primal}, and retains the optimal local coordinates $y_c^\star$. The backward call traverses the feature map and accumulates
\begin{equation}
\bar\eta
=g\sum_{c\in\mathcal C}B_c^\top y_c^\star,
\qquad
\nabla_\theta\mathcal L
=J_{\eta_\theta}(x)^\top\bar\eta
\label{eq:feature_pullback}
\end{equation}
where $\bar\eta$ is the gradient passed to the upstream coefficient model and $J_{\eta_\theta}(x)=\partial\eta_\theta(x)/\partial\theta$. Thus each stored feature reads its assigned optimal coordinate and contributes to the corresponding coefficient gradient. Automatic differentiation through the upstream model completes the update.

The conic solver supplies the optimal local coordinates, and the layer supplies their direct contraction into the coefficient gradient. No dense certificate or derivative of the full conic solution is constructed for this scalar loss. Caching the associations also avoids rebuilding the sparse template between calls. Algorithm~\ref{alg:sparse_sos_layer} gives the sequence; \Cref{tab:controlled} tests value specialization and template reuse separately.

\subsection{Cost and scope of the sparse interface}
\label{subsec:cost_scope}
Let $s_k$ be the local symmetric block sizes, $S$ the number of overlap-coupling rows, and $T$ the stored objective feature nonzeros. The block-coordinate count and storage for the cached block, overlap, and objective data are
\begin{equation}
D=\sum_k\frac{s_k(s_k+1)}{2},
\qquad
\text{storage}=\Theta(D+S+|T|)
\label{eq:representation_cost}
\end{equation}
A coefficient update scans $\Theta(|T|)$ entries and writes the $\Theta(|T_\rho|)$ entries affected by parameter $\rho$, where $T_\rho\subseteq T$. Its total cost is therefore $\Theta(|T|)$. The backward rule in \eqref{eq:feature_pullback} also takes $\Theta(|T|)$ time and constant additional working memory beyond its inputs and emitted gradient. Proposition~1 in \Cref{app:theory} gives the count-based argument. These costs exclude the conic solve and the upstream model's differentiation and explain how the interface retains the sparse representation's benefit across repeated calls.

\section{Experiments}
\label{sec:experiments}

A differentiable SOS layer must provide accurate gradients at a cost that allows the certificate to participate repeatedly in training. We evaluate \textit{Sparse SOS Layers} along these two dimensions, then examine whether optimizing the certified value improves the learned outcome. The experiments cover controlled polynomial problems, larger sparse relaxations, and learning tasks with certificate-based objectives.

Our layer is labeled \textit{Sparse SOS (Ours)} in the comparisons. We use \textit{CVXPYLayers}~\citep{agrawal2019dco,agrawal2019cone} and \textit{SDPRLayers}~\citep{holmes2025sdprlayers} as our main baselines, which we will shorten as \textit{CVXPY} and \textit{SDPR} throughout the rest of the manuscript, for simplicity. \textit{CVXPY} provides a general interface for differentiating convex optimization problems, making it a reference for the same SOS program under generic conic differentiation. \textit{SDPR} specializes in differentiable semidefinite relaxations of polynomial optimization problems, which provides a complementary reference for our Shor-relaxation tasks. Together, they allow us to compare scalar value differentiation with both general-purpose and SDP-specific approaches.

The experiments primarily answer four questions:

\begin{enumerate}
    \item \textit{Can we differentiate sparse SOS certificates accurately?}
    We first test small polynomial certificate templates where a dense reference is available.

    \item \textit{Does sparse differentiation scale?}
    We then compare against generic differentiable optimization layers on larger sparse semidefinite relaxations.

    \item \textit{Does the certificate signal change learning?}
    We test pendulum control and a design task with a required certified lower bound.

    \item \textit{Does the backward advantage survive a training loop?}
    We report a locked value-only training loop case study. 
\end{enumerate}

\subsection{Can we differentiate sparse SOS certificates accurately?}
\label{subsec:phase1}

A certificate template is a fixed polynomial expression $p_\eta(\xi) = \sum_\alpha \eta_\alpha \xi^\alpha$ whose coefficients $\eta$ come from a neural network or from a test input. The SOS layer checks whether $p_\eta$ can be represented as $p_\eta(\xi)=m(\xi)^\top Qm(\xi),~Q\succeq 0$, where $m(\xi)$ is a vector of monomials and $Q$ is a PSD Gram matrix~\citep{parrilo2003sdp,lasserre2001moments}. The sparse formulation uses smaller local matrices. The experiment asks whether the value gradient recovered from its local solver coordinates agrees with the dense reference.

We use four small sparse templates with chain, star, grid, and Erd\H{o}s random-graph interaction patterns. Each variable is a vertex, and two variables are neighbors when a local polynomial term contains both of them.

We train the scalar value interface against two baselines: (i) a \textit{stop gradient} variant that blocks the certificate gradient and (ii) a plain \textit{neural regressor} with no certificate. \textit{Stop gradient} performs the same forward solve and scalar loss but detaches the layer value, contributing zero upstream gradient. The training model is the same small multilayer perceptron in all three cases.

We check gradients against MOSEK~\citep{mosek}, a high accuracy SDP solver that we use as a numerical reference, achieving a median relative gradient error of $4.585 \times 10^{-5}$. Across all four templates, the sparse layer completes training without failures and achieves a lower final scalar loss than both the \textit{stop gradient} baseline and the plain \textit{neural regressor}.

These results show that the local solver coordinates retain the information needed for scalar value differentiation. The training comparison shows that passing this gradient to the coefficient model improves value fitting on all the tested templates.

\subsection{Does sparse differentiation scale?}
\label{subsec:scaling_families}
We next move beyond controlled templates to four larger sparse families. For a quadratic problem with variables $s$, the standard Shor relaxation~\citep{shor1987quadratic} replaces each product $s_i s_j$ with an entry of a lifted matrix $X$, imposes $X\succeq 0$ and the linear constraints implied by the original variables, and drops the nonconvex rank-one constraint $X=ss^\top$. We use this relaxation when comparing against \textit{SDPR} as it is the natural form that it operates on. For multiplier based SOS certificates, the variables are SOS multiplier polynomials and Gram matrices, with a different formulation from Shor-lifted products. In that case \textit{SDPR} is a structural mismatch, and the direct dense baseline is \textit{CVXPY} solving the same SOS SDP without sparse decomposition.

We separate controlled timing comparisons from cross-framework observations. Numerical qualification compares each analytic derivative with its own centered finite differences at three predefined step sizes. The error criterion is absolute error at most $5\times10^{-4}$ or relative error at most $5\times10^{-3}$. \Cref{tab:controlled} reports controlled timings, and \Cref{tab:large_accuracy} reports large-scale gradient errors and qualification outcomes. Small-size value and gradient checks appear in \Cref{app:scaling_quality}, cross-framework timings in \Cref{app:stack_timings}, and timing conventions in \Cref{app:experimental_protocol}.

\paragraph{Banded binary optimization.}
The task is a binary quadratic optimization problem on a ladder graph. A node is one vertex of the graph, each node carries one binary variable $s_i\in\{-1,1\}$. Edges mark pairs of variables that interact in the objective:
$\min_{s_i\in\{-1,1\}}
\sum_{(i,j)\in E} w_{ij}s_i s_j + \sum_i b_i s_i$. This is the same quadratic form used in Ising spin models and in the semidefinite relaxation behind MaxCut algorithms \citep{barahona1982computational,goemans1995improved} and the ladder graph gives the lifted matrix sparse local structure. All three methods solve this same relaxation over $N \in \{16,32,48,64,96\}$ and return strict optimal status. At $N=96$, all report objective $-95.154$, with value derivatives $0.0412$, $0.0422$, $0.0418$ for \textit{Sparse SOS}, \textit{CVXPY}, \textit{SDPR}. \Cref{app:scaling_quality} gives the complete agreement check, and \Cref{app:stack_timings} raw timings. 

We show that local solver blocks retain both the relaxation's optimal value and its sensitivity to the objective coefficients. The sparse backward pass therefore supplies a learning signal numerically consistent with the dense references on this sweep.

\paragraph{Controlled sparse baseline.}
We apply \textit{CVXPY} to the exact chordal relaxation, using $N-2$ overlapping $4\times4$ PSD blocks with duplicated overlaps constrained consistently. Holding the model, instance, parameter, and scalar loss fixed, we compare generic solution-map differentiation with direct contraction of the returned clique-local blocks. The generic route prepares its derivative operator during the forward call, so \Cref{tab:controlled} reports total warm-call time for each route. The qualified ratios are $1.62\times$, $1.95\times$, $2.21\times$, and $2.65\times$ at $N=16,32,96,128$. Thus specializing the derivative to the scalar value reduces repeated-call cost even when the generic baseline already uses the same sparse formulation. The gain comes from the derivative computation, with the solver stack and forward model kept fixed.

\begin{table}[htbp]
\centering
\small
\caption{Controlled total warm-call times, median [interquartile range] in milliseconds.}
\label{tab:controlled}
\begin{tabular}{llrr}
\toprule
Comparison & Size & Reference & Sparse value route\\
\midrule
Generic / value-specialized & $N=16$ & 28.851 [0.094] & 17.839 [0.080]\\
Generic / value-specialized & $N=128$ & 520.488 [1.550] & 196.259 [0.831]\\
Ambient / chordal & $N=16$ & 1.414 [0.003] & 3.263 [0.006]\\
Rebuilt / cached & $N=16$ & 5.893 [0.056] & 4.431 [0.048]\\
Rebuilt / cached & $N=256$ & 99.040 [5.165] & 72.820 [0.754]\\
\bottomrule
\end{tabular}
\end{table}

\paragraph{Formulation and template reuse.}
We next separate the benefit of representing the problem with local blocks from the benefit of reusing its template across calls. The ambient formulation is faster through $N=64$, the crossover occurs at $N=96$, and chordal is $4.38\times$ faster at $N=256$. This shows a size-dependent benefit from the sparse formulation even when both formulations use the same value rule.

Reusing the template gives a $1.36\times$ total-call gain at $N=256$ (\Cref{tab:controlled}). Across a $16\times$ size range, coefficient-update time grows $1.26\times$ while construction grows $14.3\times$. The comparison shows why caching matters during learning: repeated coefficient updates avoid most of the growth associated with rebuilding the model.

\paragraph{Large-scale differentiation.}
At $N=384$ on a $176$\,GiB AMD EPYC 7543 machine, the dense parameterization peaks at $164.07$\,GiB, while our layer uses $1.69$\,GiB, nearly $100\times$ less memory (\Cref{tab:large_resources}). The dense run completes one forward and one backward call before the four-hour budget stops its repeats. These cross-stack times provide resource context, without a speed ratio. The layer remains near $2$\,GiB at $N=512$, making repeated calls practical at sizes where the tested dense parameterization becomes resource-intensive.

At matched $\rho=0.45$, our derivative qualifies at all six tested sizes through $N=512$; the tested dense \textit{CVXPY} and \textit{SDPR} derivatives qualify at none of the four sizes through $N=256$ (\Cref{tab:large_accuracy}). Chordal \textit{CVXPY} qualifies at $N=96$ and $128$. At $192$, its derivative fails the error criterion at each of the three finite-difference step sizes, while the value derivative still qualifies. The checks establish qualified value differentiation at larger sizes, without diagnosing the cause of the baseline discrepancies.

\begin{table}[htbp]
\centering
\small
\caption{Resource measurements at $N=384$.}
\label{tab:large_resources}
\begin{tabular}{lrrr}
\toprule
Implementation & Peak memory (GiB) & Forward (s) & Backward (s)\\
\midrule
Dense parameterization & 164.07 & 4161 & 203.6\\
\textit{Sparse SOS (Ours)} & 1.69 & 0.154 & 0.265\\
\bottomrule
\end{tabular}
\end{table}

\begin{table}[htbp]
\centering
\small
\caption{Minimum relative derivative error over three finite-difference step sizes at $\rho=0.45$. Qualification uses the complete record and the absolute-or-relative criterion. All reported primal statuses are optimal.}
\label{tab:large_accuracy}
\begin{tabular}{rrrr}
\toprule
$N$ & \makecell{\textit{Sparse SOS}\\\textit{(Ours)}} & Dense \textit{CVXPY} & Dense \textit{SDPR}\\
\midrule
96  & $3.67\times10^{-5}$ & $2.00\times10^{-2}$ & $1.07\times10^{-2}$\\
128 & $1.88\times10^{-4}$ & $8.18\times10^{-1}$ & $5.19\times10^{-1}$\\
192 & $2.06\times10^{-5}$ & $2.46\times10^{-1}$ & $1.96\times10^{-1}$\\
256 & $3.20\times10^{-5}$ & $2.70\times10^{-2}$ & $4.06\times10^{-1}$\\
384 & $1.49\times10^{-5}$ & Not tested & Not tested\\
512 & $1.26\times10^{-5}$ & Not tested & Not tested\\
\midrule
Qualified sizes & 6/6 & 0/4 & 0/4\\
\bottomrule
\end{tabular}
\end{table}

\paragraph{Clique width and crossover.}
We evaluate $150$ independently generated random chordal instances and $50$ banded timing runs, giving $200$ qualified paired comparisons and $400$ formulation runs. Each random cell has ten independent instances and each banded cell has ten timing repetitions of one deterministic instance. \Cref{tab:width_crossover} shows where the median ambient/chordal ratio first exceeds one. Every tested width reaches a median advantage, with empirical boundary $n_{\mathrm{cross}}\approx32(w-1)$. Larger cliques delay the crossover, giving a size-width criterion for the tested families; \Cref{tab:width_full} contains the complete sweep.

\begin{table}[htbp]
\centering
\small
\caption{First tested size with median ambient/chordal time ratio above one.}
\label{tab:width_crossover}
\begin{tabular}{lrrrr}
\toprule
Family & Width $w$ & Size $n$ & Median ratio & [Q1, Q3]\\
\midrule
Banded & 4 & 96 & 1.068 & [1.063, 1.070]\\
Random chordal & 4 & 96 & 1.076 & [1.038, 1.139]\\
Random chordal & 6 & 160 & 1.057 & [1.038, 1.070]\\
Random chordal & 8 & 224 & 1.047 & [0.999, 1.104]\\
\bottomrule
\end{tabular}
\end{table}

\paragraph{Continuum robot shape projection.}
The task is to project a target shape of a serial continuum robot onto a locally feasible set. A continuum robot can be approximated by a chain of short bendable segments, and the projection asks for the closest segment configuration that respects local shape constraints. The local polynomial model gives a sparse chain structured semidefinite relaxation. \Cref{tab:robot_agreement} reports a matched chordal Shor diagnostic: relative value and gradient discrepancies are below $3.0\times10^{-3}$ with maximum clique size six. This shows that local solver coordinates recover the value sensitivity of the matched robot relaxation. \Cref{app:scaling_quality,app:robotics_chordal_shor} give the generic formulation check and full diagnostic, including solver statuses.

\begin{table}[htbp]
\centering
\small
\caption{Robot projection: relative discrepancies against \textit{SDPR} on the matched Shor relaxation.}
\label{tab:robot_agreement}
\begin{tabular}{rrr}
\toprule
Segments & Objective discrepancy & Gradient discrepancy\\
\midrule
8 & $6.80\times10^{-4}$ & $5.17\times10^{-4}$\\
9 & $2.93\times10^{-3}$ & $1.34\times10^{-4}$\\
10 & $1.50\times10^{-4}$ & $1.77\times10^{-3}$\\
11 & $1.68\times10^{-3}$ & $1.68\times10^{-3}$\\
\bottomrule
\end{tabular}
\end{table}

\paragraph{Coupled oscillator Lyapunov certificates.}
The task is to certify local Lyapunov decrease for a chain of coupled Van der Pol oscillators. A Lyapunov certificate is an energy like function that should decrease along the dynamics, if it decreases in a region it proves local stability. A Van der Pol oscillator is a standard nonlinear oscillator used as a benchmark dynamical system, and coupling several copies creates a sparse chain.

For this family, \textit{Sparse SOS Layers} and \textit{CVXPY} solve the same multiplier SOS certificate, while \textit{SDPR} is not a same-task comparator. \Cref{app:vdp_sdpr_diagnostic} explains the forced \textit{SDPR} surrogate. The sparse and dense SOS certificate values and gradients agree to displayed precision in \Cref{tab:scaling_lyapunov_quality}. The raw times for chain lengths $4$, $6$, and $8$ are reported in \Cref{tab:scaling_lyapunov}. This agreement extends the evaluation beyond quadratic Shor relaxations to sparse multiplier-based SOS certificates.

\begin{table}[htbp]
\centering
\small
\caption{Coupled oscillator certificate values and gradients for the same multiplier SOS program.}
\label{tab:scaling_lyapunov_quality}
\begin{tabular}{rcccc}
\toprule
\multirow{2}{*}{$N$} &
\multicolumn{2}{c}{Certificate value} &
\multicolumn{2}{c}{Gradient value} \\
\cmidrule(lr){2-3}\cmidrule(lr){4-5}
& \makecell{\textit{Sparse SOS}\\\textit{(Ours)}} & \textit{CVXPY} & \makecell{\textit{Sparse SOS}\\\textit{(Ours)}} & \textit{CVXPY} \\
\midrule
$4$ & $0.072584$ & $0.072584$ & $-0.242536$ & $-0.242536$ \\
$6$ & $0.108876$ & $0.108876$ & $-0.363803$ & $-0.363803$ \\
$8$ & $0.145168$ & $0.145168$ & $-0.485071$ & $-0.485071$ \\
\bottomrule
\end{tabular}
\end{table}

\paragraph{Pose graph synchronization.}
The task is angular synchronization on a sparse pose graph. Each graph node has an unknown orientation $z_i$ on the unit circle, and each edge gives a noisy relative orientation $\hat r_{ij}$. The nonconvex least squares problem is
\begin{equation}
\min_{|z_i|=1}
\sum_{(i,j)\in E} \omega_{ij}\left|z_j-\hat r_{ij}z_i\right|^2
\label{eq:pose_objective}
\end{equation}
Expanding the squared residuals gives a quadratic objective with cost matrix $C$. The semidefinite relaxation lifts $zz^\ast$ to a PSD matrix $X$, where $(\cdot)^\ast$ denotes conjugate transpose, and enforces unit diagonal constraints as
\begin{equation}
\min_{X\succeq 0} \langle C,X\rangle
\quad
\text{s.t.}
\quad
X_{ii}=1 \;\; \forall i
\label{eq:pose_sdp}
\end{equation}

\textit{SDPR} has the faster forward pass on this task, while our layer consistently has the faster backward pass (\Cref{tab:scaling_pose_graph}). The result separates forward-solver efficiency from differentiation cost as a cheaper value backward pass remains useful even when another implementation solves the forward problem faster.

\begin{table}[htbp]
\centering
\small
\caption{Pose graph synchronization: implementation-level times in seconds.}
\label{tab:scaling_pose_graph}
\begin{tabular}{rcccccc}
\toprule
Poses & \multicolumn{3}{c}{Forward time (s)} & \multicolumn{3}{c}{Backward time (s)} \\
\cmidrule(lr){2-4}\cmidrule(lr){5-7}
& \makecell{\textit{Sparse SOS}\\\textit{(Ours)}} & \textit{CVXPY} & \textit{SDPR} & \makecell{\textit{Sparse SOS}\\\textit{(Ours)}} & \textit{CVXPY} & \textit{SDPR} \\
\midrule
$10$ & $0.019$ & $0.027$ & $\mathbf{0.011}$ & $\mathbf{4.72\!\cdot\!10^{-4}}$ & $4.17\!\cdot\!10^{-3}$ & $2.57\!\cdot\!10^{-3}$ \\
$16$ & $0.023$ & $0.094$ & $\mathbf{0.016}$ & $\mathbf{1.12\!\cdot\!10^{-3}}$ & $0.026$ & $5.36\!\cdot\!10^{-3}$ \\
$22$ & $0.029$ & $0.171$ & $\mathbf{0.022}$ & $\mathbf{2.66\!\cdot\!10^{-3}}$ & $0.063$ & $9.80\!\cdot\!10^{-3}$ \\
$28$ & $0.034$ & $11.24$ & $\mathbf{0.031}$ & $\mathbf{3.41\!\cdot\!10^{-3}}$ & $0.130$ & $0.015$ \\
\bottomrule
\end{tabular}
\end{table}

\subsection{Does the certificate signal change learning?}
\label{subsec:pendulum}

The control task is to learn a stabilizing controller for a pendulum, then for chains of two and three coupled pendulums. Each pendulum contributes an angle and angular velocity to the state $x$, $u_\theta(x)$ is the learned control input and $\dot x=f_\theta(x)$ denotes the closed-loop dynamics. A Lyapunov certificate uses a nonnegative function $V(x)$ as an energy and checks its decrease on the certificate region $\mathcal R$
\begin{equation}
V(x)\geq0,\qquad
-\nabla V(x)^\top f_\theta(x)\geq\gamma,
\qquad x\in\mathcal R
\label{eq:control_certificate}
\end{equation}
where the scalar $\gamma$ is the certified margin. A positive margin means the condition holds with slack throughout $\mathcal R$, a negative margin means the certificate failed. The controller trained with \textit{Sparse SOS} receives the gradient of this margin during training. The \textit{sampled surrogate} uses a grid-based loss, while \textit{post hoc} certification evaluates the controller only after training.

\Cref{tab:pendulum} reports the same certificate metrics for all three systems. In the single-pendulum case, both \textit{Sparse SOS} and \textit{sampled surrogate} certify. An additional check on ten seeds also gives certification rate $1.00$ for both, with a margin difference of $0.004$-$0.005$. On the coupled two- and three-pendulum systems, all main methods certify and the margin differences are approximately $10^{-3}$. Single- and two-pendulum results are means with standard errors over five seeds while the three-pendulum result is a short validation.

\begin{table}[htbp]
\centering
\small
\caption{Pendulum control certificate outcomes. }
\label{tab:pendulum}
\begin{tabular}{lllrr}
\toprule
Pendulum & Method & Training signal & Cert.\ margin & Cert.\ rate \\
\midrule
1 & \textit{Sparse SOS (Ours)} & certificate margin & $\mathbf{0.0274 \pm 0.0014}$ & 1.000 \\
1 & \textit{Sampled surrogate} & sampled grid loss & $0.0234 \pm 0.0013$ & 1.000 \\
1 & \textit{Post hoc} & no certificate during training & $-0.7957 \pm 0.7453$ & 0.200 \\
\midrule
2 & \textit{Sparse SOS (Ours)} & certificate margin & $\mathbf{0.0036 \pm 0.0002}$ & 1.000 \\
2 & \textit{Sampled surrogate} & sampled grid loss & $0.0024 \pm 0.0002$ & 1.000 \\
2 & \textit{Post hoc} & no certificate during training & $0.0019 \pm 0.0004$ & 1.000 \\
\midrule
3 & \makecell[l]{\textit{Sparse SOS (Ours)}\\every step} & certificate margin & $\mathbf{0.0021}$ & 1.000 \\
3 & \textit{Sampled surrogate} & sampled grid loss & $0.0014$ & 1.000 \\
3 & \textit{Post hoc} & no certificate during training & $0.0014$ & 1.000 \\
\bottomrule
\end{tabular}%
\end{table}

The control results show that the scalar certificate can participate directly in controller training and improve the certified margin. The gain is modest in these small systems, where the \textit{sampled surrogate} also certifies successfully. Additional diagnostics are given in \Cref{app:pendulum_ablations,app:multi_pendulum,app:cert_native}.

\paragraph{Learning a design with a certified requirement.}
To test a consequential learning outcome, an MLP predicts a design whose certified lower bound must improve over nominal by $\delta=0.005$. Training uses $n\in\{24,48\}$. Evaluation consist of test problem instances including the larger sizes $n=96,160$, which are both absent from training. \Cref{tab:design} pools violations over the five seeds. Denominators count evaluated designs, with the same test instances reconstructed for each seed. The final column counts training optimization solves per seed.

\textit{Sparse SOS} uses the proposed layer to differentiate the certified value. The \textit{finite differences} estimates that gradient through perturbed optimization solves, while the \textit{sampled surrogate} uses a sample-based training objective. \textit{Stop gradient} retains the certificate solve but blocks its derivative, \textit{no layer} trains without the certificate layer. 

\begin{table}[htbp]
\centering
\small
\caption{Violations of the certified lower-bound requirement.}
\label{tab:design}
\begin{tabular}{lrrrrr}
\toprule
Training signal & Train & Test & $n=96$ & $n=160$ & Solves\\
\midrule
\textit{Sparse SOS (Ours)} & 0/80 & 0/30 & 0/20 & 0/30 & 480\\
\textit{Finite differences} & 4/80 & 0/30 & 4/20 & 6/30 & 3360\\
\textit{Sampled surrogate} & 9/80 & 4/30 & 5/20 & 12/30 & 0\\
\textit{Stop gradient} & 80/80 & 30/30 & 20/20 & 30/30 & 480\\
\textit{No layer} & 80/80 & 30/30 & 20/20 & 30/30 & 0\\
\bottomrule
\end{tabular}
\end{table}

Results shown in (\Cref{tab:design}). \textit{Sparse SOS} is the only method with zero violations on every split and seed, using 1/7 of the training optimization solves of \textit{finite differences}. At $n=160$, its worst design margin is $0.0056$, just above the required $0.005$. The \textit{sampled surrogate} succeeds on three of five seeds, so exact differentiation is not necessary in every run. Failures cluster within seeds; the $30$ evaluated designs are not $30$ independent training outcomes. \Cref{app:design_outcomes} gives the per-seed counts and margins. The result demonstrates a consequence of the scalar gradient beyond value fitting: it changes the learned designs so that all evaluated designs meet the certified requirement, with fewer training solves than \textit{finite differences}.

\subsection{Does the backward advantage survive a training loop?}
\label{subsec:training_loop}

The previous benchmarks show that value differentiation can reduce the cost of repeated calls. This case study evaluates end-to-end training when every gradient step differentiates through an optimization layer.

The task is the banded binary optimization problem at $N=96$. Each training example is a noisy ladder graph with a feature vector $h$. A small shared model predicts three edge class weights
$\alpha_\theta(h) =
(\alpha_{\mathrm{nearest}},\alpha_{\mathrm{next}},\alpha_{\mathrm{field}})$,
which reweight the objective matrix of the same Shor semidefinite relaxation used in \Cref{tab:scaling_combinatorial}. The layer solves that relaxation and returns the scalar value $v(\alpha_\theta(h))$. Each instance has a precomputed target value $t$, and all methods minimize
\begin{equation}
\ell(\theta;h,t)=\left(v(\alpha_\theta(h))-t\right)^2
\label{eq:training_loss}
\end{equation}
In the experiments, we only vary the differentiable layer used to solve and differentiate through the relaxation. The exact model, features, split, and label generation are described in \Cref{app:q3_training_details}.

The results are presented in \Cref{tab:training_loop}. \textit{Sparse SOS} reaches the lowest final loss and the normalized loss threshold $10^{-3}$ faster than both baselines. The cumulative times reflect complete implementations with different numerical solver stacks, complementing the isolated comparison in \Cref{tab:controlled}. 


\begin{table}[htbp]
\centering
\small
\caption{Training loop case study at problem size $96$.}
\label{tab:training_loop}
\begin{tabular}{lrrrr}
\toprule
Layer choice & Final test loss & Forward total (s) & Backward total (s) & Time-to-$10^{-3}$ (s)\\
\midrule
\textit{Sparse SOS (Ours)} & $\mathbf{1.06\!\cdot\!10^{-7}}$ & 32.93 & 5.37 & 22.28\\
\textit{SDPR} & $3.78\!\cdot\!10^{-5}$ & 20.67 & 268.33 & 71.94\\
\textit{CVXPY} & $3.66\!\cdot\!10^{-5}$ & 2023.60 & 361.12 & 594.36\\
\bottomrule
\end{tabular}
\end{table}

The sparse value pullback reduces cumulative backward time inside this locked same-task training loop. The lower time to the common loss threshold shows that the computational benefit survives repeated parameter updates and reaches a learning outcome sooner in this experiment.

\section{Conclusion}
\label{sec:conclusion}

\textit{Sparse SOS Layers} makes certified scalar values practical training objectives for fixed sparse SOS programs. The experiments show that specializing differentiation to the optimal value reduces repeated-call cost, while preserving local solver coordinates keeps memory use manageable at sizes where the tested dense parameterization becomes expensive. In learning, this gradient improves certified design outcomes and reduces the optimization solves needed compared with \textit{finite differences}.

These findings support learning through an existing sparse verifier when the learned quantities enter its objective and its template remains fixed. The useful regime depends on clique width and problem size, and the guarantees remain those of the chosen relaxation. A separate limitations discussion in \Cref{sec:limitations} details these scope and numerical dependencies.

Together, these results offer a practical route to certificate-based learning in control, robotics, and structured optimization with fixed sparse verifiers. Future work will evaluate larger coefficient models and longer training runs to establish how these repeated-call gains translate to more demanding learning tasks.

\label{maintext:end}
\clearpage
\begin{ack}

This work is supported by Defi Clé Robotique centrée sur l’Humain funded by Région Occitanie, France and was granted access to the HPC resources of IDRIS under the allocation AD011017069 made by GENCI, and was also funded by ANITI IA Cluster (ANR-23-IACL-0002) and the HAMMER project of the PEPR Acceleration Robotics program (ANR-25-PERO-0002).

\end{ack}

\bibliographystyle{unsrtnat}
\bibliography{ref}

\clearpage
\appendix
\section{Formal Statements}
\label{app:theory}

This appendix collects the formal statements used by the layer. The principles are standard sensitivity, perturbation, and value-function facts \citep{fiacco1983sensitivity,bonnans2000perturbation,mehmood2020valuefunction}. The point is to apply them to the sparse representation actually used by the SOS certificate. \textit{Lemma 1} and \textit{Theorem 1} support the scalar value pullback, \textit{Proposition 1} counts the operations used by the fixed-template architecture.

\paragraph{Assumption Set A: Value-only layer.}
At a reference point $\theta_0$, assume: (i) sparse SOS relaxation data are fixed in a neighborhood of $\theta_0$, (ii) exposed coefficient map $c(\theta)$ is continuously differentiable, (iii) feasible set $Y$ is nonempty, closed, and independent of $\theta$ in that neighborhood, (iv) the optimal solution sets are nonempty and contained in a common compact set for $\theta$ near $\theta_0$, (v) strong duality holds, (vi) either the optimizer is locally unique, or a subgradient statement is acceptable, (vii) sparse moment coordinates used by the implementation are locally identifiable from the active sparse basis.

\subsection*{Lemma 1: Envelope reduction for the value map}
Let
$v(\theta) = \min_{y \in Y} \langle c(\theta), y \rangle$,
under \textit{Assumption Set A}. Then for every direction $h$,
$v'(\theta_0;h)=\min_{y \in Y^\star(\theta_0)} \langle J_c(\theta_0)h, y \rangle$.
If $Y^\star(\theta_0)=\{y^\star\}$, then $v$ is differentiable at $\theta_0$ and
$\nabla v(\theta_0)=J_c(\theta_0)^\top y^\star$.

\textit{Proof:}
Because the feasible set is fixed locally, the relevant optimal set is locally bounded, and $\theta \mapsto \langle c(\theta), y \rangle$ is $C^1$ for each feasible $y$, $v$ is a pointwise minimum of smooth functions over a common feasible set with locally compact argmin sets. Danskin's theorem yields the directional derivative formula. If the optimizer is unique, the directional derivative is linear in $h$, so the gradient formula follows. If the optimizer is not unique, the same formula yields the Clarke subgradient set generated by $J_c(\theta_0)^\top y$ over optimal $y$. \hfill $\square$

\subsection*{Theorem 1: Value-only sparse pullback}
Let the layer return the certified scalar
$v(\theta)=\min_{y\in Y}\langle c(\theta),y\rangle$ under \textit{Assumption Set A}. For an upstream scalar adjoint $g=\partial \mathcal L/\partial v$, the reverse-mode pullback at a locally unique optimizer $y^\star$ is
$\nabla_\theta \mathcal L = g\,J_c(\theta_0)^\top y^\star$.
If the optimizer is not unique, the same expression with any returned optimal sparse coordinate gives the corresponding subgradient element.

\textit{Proof:}
Apply \textit{Lemma 1} and multiply the value derivative by the upstream adjoint $g$. The implementation-specific identifiability assumption ensures that the sparse moment vector recovered by the layer is exactly the optimal coordinate vector needed in this pullback. \hfill $\square$

\subsection*{Proposition 1: Fixed-template representation and pullback cost}
Let the local symmetric blocks have sizes $s_k$, let $S$ count the overlap-coupling rows, and let $T$ contain the stored objective feature nonzeros. Define $D=\sum_k s_k(s_k+1)/2$. The block, overlap, and objective representation requires $\Theta(D+S+|T|)$ storage. A parameter update scans $\Theta(|T|)$ entries and writes $\Theta(|T_\rho|)$ affected coefficients, where $T_\rho\subseteq T$, so its total cost is $\Theta(|T|)$. Returned solver coordinates occupy $\Theta(D)$ storage. The pullback takes $\Theta(|T|)$ time and $\Theta(1)$ working memory beyond the incoming adjoint, coordinates read, and emitted gradient.

\textit{Proof:}
Each symmetric block contributes $s_k(s_k+1)/2$ coordinates. Overlap rows store the equalities between duplicated coordinates, and the objective map stores one record per feature nonzero. Their counts add to $D+S+|T|$, any additional fixed affine constraints have their own sparse storage. The update visits every stored feature and writes each affected coefficient. Since $T_\rho\subseteq T$, scan and write work is $\Theta(|T|)$. The reverse pass reads the assigned local coordinate for each feature and accumulates its coefficient derivative in \eqref{eq:value_pullback}. A sequential traversal uses constant working memory outside the stored inputs and output. These counts exclude the conic solve and the upstream model's own differentiation cost. \hfill $\square$
\section{Practical Notes}
\label{app:algorithm}
\subsection{Implementation and timing conventions}
\label{app:experimental_protocol}
In our implementation the sparse pattern, monomial basis, and clique layout are fixed once the task is chosen. The layer caches basis maps, sparsity metadata, overlap constraints, and the objective feature map. Across repeated calls, only the learned objective coefficients are updated. The forward sparse SOS solve is built on top of an existing engine that supports correlative and term sparsity \citep{waki2006sparse,wang2021tssos}.

The controlled comparisons use one fixed instance per cell, five untimed warmups, twenty retained samples, and summaries of the median, interquartile range (IQR), minimum, and maximum. Construction is separated from warm calls, and sub-millisecond measurements are batched. Qualification compares each analytic derivative with its own centered finite differences at three predeclared steps using absolute error at most $5\times10^{-4}$ or relative error at most $5\times10^{-3}$. The generic \textit{CVXPY} route prepares its derivative operator during the forward call, so the specialization comparison uses total warm-call time for each route.

Each comparison in \Cref{tab:controlled} changes one factor. Value specialization uses the same chordal model, formulation uses the same Julia/JuMP/MOSEK stack and value rule, template reuse uses the same model and solver.

The large-resource study uses a $176$\,GiB AMD EPYC 7543 machine. At $N=384$, dense canonicalization constructs an \texttt{int64} index array of $2.20\times10^{10}$ entries, approximately $164$\,GiB before other arrays. The individual completed calls are reported separately from repeated timing distributions. Other measurements use a single 20-thread Intel i9-9900X CPU. Framework times include canonicalization, caching, solver-interface overhead, and the forward solver selected by each implementation.

\subsection{Layer Algorithm}
Algorithm~\ref{alg:sparse_sos_layer} retains the optimal sparse coordinates needed by the scalar value pullback.
\begin{algorithm}[htbp]
\small
\caption{Forward and backward pass of the sparse SOS value layer.}
\label{alg:sparse_sos_layer}
\begin{algorithmic}[1]
\Require Fixed basis, relaxation order, cliques, constraints, and coefficient maps $B_c$
\State Construct and cache the sparse conic model, feature map, and local-coordinate references
\Require Input $x$, model parameters $\theta$, upstream adjoint $g=\partial\mathcal L/\partial v$
\State Compute exposed coefficients $\eta_\theta(x)$ from the upstream model
\State Scan the feature map to update $c_c=c_c^0+B_c\eta_\theta(x)$
\State Solve the sparse conic program in \eqref{eq:sparse_sos_primal}
\State Return $v_\theta(x)$ and retain optimal local coordinates $y_c^\star$
\State Accumulate $\bar\eta=g\sum_c B_c^\top y_c^\star$ using the same sparse feature map
\State Pass $\bar\eta$ to the upstream model to obtain $\nabla_\theta\mathcal L$ as in \eqref{eq:feature_pullback}
\end{algorithmic}
\end{algorithm}

\section{Limitations}
\label{sec:limitations}

The implemented interface supports learned objective coefficients under a fixed basis, relaxation order, constraints, localizing polynomials, sparsity pattern, and clique structure. These conditions let the layer reuse its template and apply the value derivative. Learned operators, region radii, and changing sparsity remain unsupported. Losses on individual certificate matrices also fall outside the scalar-value interface.

The certificate and its gradient describe the chosen conic relaxation up to solver tolerance. A weak relaxation or unsuitable certificate template can therefore limit what training achieves, even when its value derivative is accurate. At degenerate optima, the value can have several sensitivities, and the implementation uses the optimal coordinates selected by the solver. The numerical checks in \Cref{tab:large_accuracy,app:scaling_quality} test the reported instances.

Efficiency depends on problem size, clique width, and repeated use of a fixed template. Larger cliques delay the measured crossover, and our layer need not have the fastest forward solver, as the pose graph results show. Controlled comparisons isolate the value rule, formulation, and template reuse. Framework timings also include differences in software and solvers and cannot attribute the full time gap to the backward rule. The oscillator comparison is limited to the two implementations solving the same multiplier SOS certificate, the forced alternative in \Cref{app:vdp_sdpr_diagnostic} solves a different problem.

The learning experiments use small models and structured, generated problems. The design study's zero-violation result applies to its evaluated instances, sizes, and seeds, with some margins close to the requirement. The \textit{sampled surrogate} also satisfies that requirement in three seeds, and both the \textit{sampled surrogate} and \textit{Sparse SOS} certify the small pendulum systems. These results establish benefits for the tested value-driven tasks. Broader learning applications require evaluation with their own certificate templates and data.

\section{Additional Experiments}
\label{app:tables}

\subsection{Objective and gradient quality checks for scaling}
\label{app:scaling_quality}

Runtime tables are meaningful only if the compared layers solve the same relaxation. The following tables report the scalar values used to check this. For Shor-relaxation tasks, objective value is the optimum of the lifted semidefinite relaxation, and gradient value is the derivative of that optimum with respect to the scalar benchmark parameter. For the oscillator Lyapunov task, certificate value is the certified SOS margin, and gradient value is the directional derivative used in the parameter-update benchmark.

\Cref{tab:scaling_combinatorial_quality} shows the results. Objectives agree to displayed precision at all sizes, and the gradients remain close even at the largest instance. These checks establish same-relaxation agreement for this small-size benchmark, they are distinct from the matched-parameter qualification study in \Cref{tab:large_accuracy}.

Pose graph synchronization gives another direct Shor comparison (\Cref{tab:scaling_pose_graph_quality}). The objective and gradient values match to numerical precision, which supports using that task as a same-relaxation comparison.

The oscillator values and gradients in \Cref{tab:scaling_lyapunov_quality} compare the two methods solving the same multiplier based SOS certificate. Their agreement supports the same-certificate timing comparison in \Cref{tab:scaling_lyapunov}, although the implementation stacks differ.

\Cref{tab:scaling_robotics_quality} reports the continuum robot check. The values are close, but not at the same numerical level as the banded binary and pose graph tasks. This is the reason the robot row is treated as a stress test for the generic sparse SOS implementation. The chordal Shor diagnostic in \Cref{app:robotics_chordal_shor} isolates the representation issue more cleanly.

\begin{table}[H]
\centering
\small
\caption{Banded binary optimization quality check.}
\label{tab:scaling_combinatorial_quality}
{%
\begin{tabular}{rcccccc}
\toprule
\multirow{2}{*}{$N$} &
\multicolumn{3}{c}{Objective value} &
\multicolumn{3}{c}{Gradient value} \\
\cmidrule(lr){2-4}\cmidrule(lr){5-7}
& \makecell{\textit{Sparse SOS}\\\textit{(Ours)}} & \textit{CVXPY} & \textit{SDPR} & \makecell{\textit{Sparse SOS}\\\textit{(Ours)}} & \textit{CVXPY} & \textit{SDPR} \\
\midrule
$16$ & $-15.213$ & $-15.213$ & $-15.213$ & $-0.0131$ & $-0.0131$ & $-0.0131$ \\
$32$ & $-31.424$ & $-31.424$ & $-31.424$ & $-0.0209$ & $-0.0209$ & $-0.0209$ \\
$48$ & $-47.367$ & $-47.367$ & $-47.367$ & $-5.575\!\cdot\!10^{-3}$ & $-5.596\!\cdot\!10^{-3}$ & $-5.569\!\cdot\!10^{-3}$ \\
$64$ & $-63.118$ & $-63.118$ & $-63.118$ & $0.0197$ & $0.0198$ & $0.0197$ \\
$96$ & $-95.154$ & $-95.154$ & $-95.154$ & $0.0412$ & $0.0422$ & $0.0418$ \\
\bottomrule
\end{tabular}%
}

\end{table}

\begin{table}[H]
\centering
\small
\caption{Continuum robot shape projection quality check for the generic sparse SOS implementation.}
\label{tab:scaling_robotics_quality}
{%
\begin{tabular}{rcccccc}
\toprule
\multirow{2}{*}{Segments} &
\multicolumn{3}{c}{Objective value} &
\multicolumn{3}{c}{Gradient value} \\
\cmidrule(lr){2-4}\cmidrule(lr){5-7}
& \makecell{\textit{Sparse SOS}\\\textit{(Ours)}} & \textit{CVXPY} & \textit{SDPR} & \makecell{\textit{Sparse SOS}\\\textit{(Ours)}} & \textit{CVXPY} & \textit{SDPR} \\
\midrule
$8$ & $1.322\!\cdot\!10^{-3}$ & $1.367\!\cdot\!10^{-3}$ & $1.299\!\cdot\!10^{-3}$ & $3.235\!\cdot\!10^{-3}$ & $3.498\!\cdot\!10^{-3}$ & $3.206\!\cdot\!10^{-3}$ \\
$9$ & $7.619\!\cdot\!10^{-4}$ & $7.649\!\cdot\!10^{-4}$ & $7.451\!\cdot\!10^{-4}$ & $2.931\!\cdot\!10^{-3}$ & $3.002\!\cdot\!10^{-3}$ & $2.882\!\cdot\!10^{-3}$ \\
$10$ & $7.818\!\cdot\!10^{-4}$ & $7.897\!\cdot\!10^{-4}$ & $7.592\!\cdot\!10^{-4}$ & $3.074\!\cdot\!10^{-3}$ & $3.245\!\cdot\!10^{-3}$ & $2.990\!\cdot\!10^{-3}$ \\
$11$ & $7.959\!\cdot\!10^{-4}$ & $8.026\!\cdot\!10^{-4}$ & $7.702\!\cdot\!10^{-4}$ & $3.085\!\cdot\!10^{-3}$ & $3.217\!\cdot\!10^{-3}$ & $2.976\!\cdot\!10^{-3}$ \\
\bottomrule
\end{tabular}%
}

\end{table}

\begin{table}[H]
\centering
\small
\caption{Pose graph synchronization quality check.}
\label{tab:scaling_pose_graph_quality}
{%
\begin{tabular}{rcccccc}
\toprule
\multirow{2}{*}{Poses} &
\multicolumn{3}{c}{Objective value} &
\multicolumn{3}{c}{Gradient value} \\
\cmidrule(lr){2-4}\cmidrule(lr){5-7}
& \makecell{\textit{Sparse SOS}\\\textit{(Ours)}} & \textit{CVXPY} & \textit{SDPR} & \makecell{\textit{Sparse SOS}\\\textit{(Ours)}} & \textit{CVXPY} & \textit{SDPR} \\
\midrule
$10$ & $-0.253278$ & $-0.253278$ & $-0.253278$ & $-5.197106$ & $-5.197109$ & $-5.197106$ \\
$16$ & $-0.401243$ & $-0.401243$ & $-0.401243$ & $-8.269580$ & $-8.269584$ & $-8.269580$ \\
$22$ & $-0.543913$ & $-0.543927$ & $-0.543914$ & $-11.201$ & $-11.201$ & $-11.201$ \\
$28$ & $-0.694724$ & $-0.694726$ & $-0.694726$ & $-14.299$ & $-14.299$ & $-14.299$ \\
\bottomrule
\end{tabular}%
}

\end{table}

\paragraph{Polynomial-chain derivative check.}
For the chain in \Cref{subsec:local_blocks}, the smallest supported instance from our benchmark generator has $n=3$ and uses $(a_1,a_2,a_3)=(-0.036,-0.064,-0.069)$, $b_{12}=0.500$, and $b_{23}=0.359$. The solve returns optimal status and certified value $0.9220524061$. Both blocks agree on $L(\xi_2)=-0.40588$ and $L(\xi_2^2)=0.16474$. The derivative with respect to the $\xi_1\xi_2$ coefficient is the returned moment $L(\xi_1\xi_2)=-0.15874640$. Centered \textit{finite differences} give $-0.15874709$ at step $10^{-2}$ and $-0.15874733$ at $10^{-3}$, agreeing to $9.3\times10^{-7}$. This check connects a local solver coordinate to the value derivative without a dense reconstruction.

\subsection{Exact chordal Shor diagnostic for the robot task}
\label{app:robotics_chordal_shor}

The continuum robot task is quadratic after the trigonometric terms are represented by lifted sine and cosine variables. The generic sparse SOS implementation used for the robot scaling row exploits correlative sparsity automatically. That is the relevant implementation path for a general sparse certificate stack, but it can introduce a sparse representation that is not numerically identical to the dense lifted matrix used by \textit{SDPR}.

For this diagnostic, we used a stricter formulation. We first formed the exact lifted Shor matrix support required by the robot objective and equality constraints. We then chordally completed that support and introduced one PSD matrix for each maximal clique. Equality constraints tie entries that appear in more than one clique. By the PSD completion theorem for chordal graphs~\citep{grone1984pdcompletion}, this represents the same sparse PSD completion condition as the dense lifted Shor matrix on the specified support. The scalar backward pass uses the envelope derivative from the solved clique moment entries and does not reconstruct a dense certificate matrix.

\Cref{tab:robotics_chordal_shor_diagnostic} demonstrates that when the same Shor relaxation is exposed through chordal PSD completion cliques, sparse value differentiation recovers close agreement and remains efficient, where the agreement is relative to \textit{SDPR} on the same lifted Shor relaxation. Relative objective and gradient discrepancies versus \textit{SDPR} are below $3.0\times 10^{-3}$, the largest clique has size six, and the scalar pullback uses the returned clique coordinates.

\begin{table}[H]
\centering
\small
\caption{Exact chordal Shor diagnostic for continuum robot projection.}
\label{tab:robotics_chordal_shor_diagnostic}
{%
\begin{tabular}{rcccccccc}
\toprule
\multirow{2}{*}{Seg.} &
\multicolumn{2}{c}{Status rate} &
\multicolumn{2}{c}{Agreement vs. \textit{SDPR}} &
\multicolumn{2}{c}{Time (s)} &
\multicolumn{2}{c}{Chordal structure} \\
\cmidrule(lr){2-3}\cmidrule(lr){4-5}\cmidrule(lr){6-7}\cmidrule(lr){8-9}
& Usable & Strict & Obj. rel. & Grad. rel. & Forward & Backward & Cliques & Max clique \\
\midrule
$8$  & $1.0$ & $1.0$ & $6.80\!\cdot\!10^{-4}$ & $5.17\!\cdot\!10^{-4}$ & $2.99\!\cdot\!10^{-2}$ & $6.67\!\cdot\!10^{-3}$ & $34$ & $6$ \\
$9$  & $1.0$ & $1.0$ & $2.93\!\cdot\!10^{-3}$ & $1.34\!\cdot\!10^{-4}$ & $3.20\!\cdot\!10^{-2}$ & $7.58\!\cdot\!10^{-3}$ & $38$ & $6$ \\
$10$ & $1.0$ & $0.8$ & $1.50\!\cdot\!10^{-4}$ & $1.77\!\cdot\!10^{-3}$ & $3.57\!\cdot\!10^{-2}$ & $8.94\!\cdot\!10^{-3}$ & $42$ & $6$ \\
$11$ & $1.0$ & $0.8$ & $1.68\!\cdot\!10^{-3}$ & $1.68\!\cdot\!10^{-3}$ & $4.20\!\cdot\!10^{-2}$ & $1.02\!\cdot\!10^{-2}$ & $46$ & $6$ \\
\bottomrule
\end{tabular}%
}
\end{table}

\subsection{Validation of SOS certificates}
\label{app:independent_checker}

Independent validation provides an additional check for errors in the control experiments. It takes saved certificate candidates, reconstructs the polynomial inequalities, and evaluates the same mathematical sign condition on test, infeasible, and near-boundary stress cases. Across the audit, the checker accepts all $18/18$ test certified cases and finds zero false certifications among $36$ rejected or stress cases. The checker margins are $0.526$ for test cases, $-0.017$ for infeasible cases, and $-0.001$ for stress cases, where positive values indicate acceptance and negative values indicate rejection. The audit-relevant quantity is the categorical pass/fail agreement, as the independent reconstruction accepts the feasible cases and rejects the designed failures.

\subsection{Forced \textit{SDPR} diagnostic for the oscillator certificate}
\label{app:vdp_sdpr_diagnostic}

The coupled oscillator experiment is a multiplier based SOS Lyapunov certificate. The certificate asks for polynomial multipliers that prove a local inequality of the form ``the Lyapunov function decreases on the certified region.'' \textit{Sparse SOS} and \textit{CVXPY} solve this same order two Putinar certificate SDP. However, our method keeps sparse Gram blocks explicit, while \textit{CVXPY} uses a dense generic representation.

\textit{SDPR} is not a direct comparator for this certificate because its natural interface is a Shor relaxation~\citep{shor1987quadratic} of a low order polynomial optimization problem. A Shor relaxation starts from a quadratic objective and quadratic constraints over decision variables, then lifts products of variables into a PSD matrix. The oscillator certificate's decision variables are SOS multiplier polynomials and Gram matrices that certify a polynomial inequality. There is no single low order QCQP whose Shor relaxation is exactly this multiplier certificate.

For diagnostic purposes, we tested the closest \textit{SDPR} style surrogate, which is a degree two monomial lift with consistency constraints and midpoint shell equalities for the oscillator variables. This surrogate uses the same dynamics parameters, but it no longer returns the certified Lyapunov margin from the SOS multiplier problem. It solves successfully at the tested chain lengths, but its strict or rank tight rate is zero and its reported value is not the certificate margin. This demonstrates a structural mismatch, \textit{SDPR} can be forced onto a nearby lifted problem, but that problem is not the certificate solved by the \textit{Sparse SOS} and \textit{CVXPY} rows.

\subsection{Value-learning case study details}
\label{app:q3_training_details}

The value-learning case study uses the banded binary family at $N=96$. The dataset has $12$ training, $4$ validation, and $8$ test instances for one locked seed. Each instance has a clean amplitude and clean coupling parameter used to generate the scalar target value, plus a noisy amplitude and noisy coupling parameter exposed to the learner. The scalar target is the exact combinatorial optimum of the clean ladder graph computed by dynamic programming, normalized by $N$.

The feature vector is $h=[1,\widetilde a,\widetilde \rho,|\widetilde \rho|,N/100]$, where $\widetilde a$ and $\widetilde \rho$ are the noisy amplitude and noisy coupling parameter. The model is a single affine map from these five features to three edge-class logits, followed by a positive softplus transform $\alpha_\theta(h)=0.05+\operatorname{softplus}(Wh+b)\in\mathbb R_{>0}^3$. The three outputs multiply the nearest-neighbor, next-nearest-neighbor, and field parts of the objective matrix. The same model initialization, Adam optimizer, learning rate $0.02$, eight epochs, split, labels, and Shor relaxation are used for \textit{Sparse SOS}, \textit{SDPR}, and \textit{CVXPY}. Wall-clock-to-threshold is the first cumulative training time at which the normalized scalar-value loss is at most $10^{-3}$.

The sparse SOS row uses the value-only envelope pullback with cached support readout, it does not use finite differences. This demonstrates that the training-loop result is a same-task learning comparison. The model, scalar labels, optimizer family, and semidefinite relaxation are fixed, while the differentiable layer implementation changes.

\subsection{Single pendulum ablations and generalization}
\label{app:pendulum_ablations}

The single pendulum diagnostics ask whether the certificate outcome is stable under distribution shift and update schedules. \Cref{tab:pendulum_single_full} keeps the timing columns for the three main training modes. \Cref{tab:generalization} shows that the controller trained with \textit{Sparse SOS} retains positive margins under several shifts, with the largest gains over \textit{stop gradient} certification on the larger-region and combined-hard settings.

Replacing the implicit gradient with a \textit{stop gradient} drops the certified margin from $0.028$ to $-0.323$, showing that this controller benefits from the certificate gradient in that comparison. \Cref{tab:schedule_ablation} shows that using exact certificate solves less frequently can reduce training time while preserving a positive margin. 

\textit{Every step} evaluates the certificate loss at each training step, \textit{every 5} and \textit{every 10} evaluate it periodically, \textit{late-stage} uses cheap training first and certificate updates only near the end, and \textit{curriculum} uses a warm start before an exact-certificate phase. Reported exact-step fraction counts the certificate-training phase reported by the script, for curriculum rows it does not count the warm-start phase. Minor model and basis variants kept the certified margin in $[0.023,0.028]$, thus they do not alter the conclusion. Together, these diagnostics describe the behavior of the scalar certificate signal in this small control problem.

\begin{table}[H]
\centering
\small
\caption{Single pendulum control full table.}
\label{tab:pendulum_single_full}
{%
\begin{tabular}{llrrrrr}
\toprule
Method & Training signal & Cert.\ margin & Cert.\ rate & \makecell{Train\\time (s)} & \makecell{Train speedup\\vs layer} \\
\midrule
\textit{Sparse SOS (Ours)} & certificate margin  & $0.0274 \pm 0.0014$ & 1.000 & 52.86 & $1.0\times$ \\
\textit{Sampled surrogate} & sampled grid loss  & $0.0234 \pm 0.0013$ & 1.000 & 27.97 & $1.9\times$ \\
\textit{Post hoc} & -  & $-0.7957 \pm 0.7453$ & 0.200 & 11.38 & $4.6\times$ \\
\bottomrule
\end{tabular}%
}

\end{table}

\begin{table}[H]
\centering
\small
\caption{Pendulum generalization under distribution shift. $\Delta$ is layer $-$ baseline certified margin.}
\label{tab:generalization}
\begin{tabular}{lccccc}
\toprule
Distribution shift & Layer margin & Layer success &
\makecell{$\Delta$ margin vs sampled} &
\makecell{$\Delta$ margin vs stop-grad} \\
\midrule
in-distribution    & 0.028 & 1.000 & 0.004 & 0.351 \\
high gravity       & 0.025 & 1.000 & 0.002 & 0.223 \\
low damping        & 0.024 & 1.000 & 0.003 & 0.349 \\
large region       & 0.054 & 0.970 & 0.008 & 0.836 \\
combined hard      & 0.050 & 0.957 & $5.387\!\cdot\!10^{-4}$ & 0.634 \\
\bottomrule
\end{tabular}

\end{table}

\begin{table}[H]
\centering
\small
\caption{Pendulum schedule ablation.}
\label{tab:schedule_ablation}
{%
\begin{tabular}{lrrrrr}
\toprule
Schedule & Cert margin & Train time (s) & \makecell{Train speedup\\vs every step}  & Cert time (s) & Exact step frac \\
\midrule
\textit{every step} & 0.028 & 44.942 & $1.0\times$ & 0.019 & 1.000 \\
\textit{every 5} & 0.024 & 20.235 & $2.2\times$ & 0.008 & 0.202 \\
\textit{every 10} & 0.023 & 18.804 & $2.4\times$ & 0.007 & 0.101 \\
\textit{late-stage} & 0.025 & 21.495 & $2.1\times$ & 0.007 & 0.183 \\
\textit{curriculum} & 0.028 & 22.115 & $2.0\times$ & 0.007 & 1.000 \\
\bottomrule
\end{tabular}%
}

\end{table}

\subsection{Coupled pendulum verifications}
\label{app:multi_pendulum}

The coupled-pendulum diagnostics extend the control check to sparse interacting systems. \Cref{tab:coupled_pendulum_full_diagnostics} adds \textit{stop gradient}, \textit{DSOS}, and \textit{SDSOS} rows to the two pendulum chain. All methods certify at this size, so the result is not a binary success/failure separation. Instead, it demonstrates that exact in-loop certificate feedback gives the largest margin within the matched controller family.

\Cref{tab:coupled_pendulum_schedule} shows that the two pendulum chain admits cheaper schedules with some margin loss. The every-step mean includes one cold-start outlier; the schedule comparison is therefore useful as an implementation diagnostic, not as a statistical claim about exact wall-clock ratios. \Cref{tab:three_pendulum} is a short budget validation at three pendula. It demonstrates that the implementation still runs and certifies on a larger coupled chain, but it should not be read as a full scaling study.

\begin{table}[H]
\centering
\small
\caption{Coupled two pendulum chain full diagnostic matrix.}
\label{tab:coupled_pendulum_full_diagnostics}
{%
\setlength{\tabcolsep}{4pt}
\begin{tabular}{llrrrrr}
\toprule
Method & Training signal & Cert.\ margin & Cert.\ rate & \makecell{Train\\time (s)} & \makecell{Train speedup\\vs layer} \\
\midrule
\textit{Sparse SOS (Ours)} & certificate margin  & $0.0036 \pm 0.0002$ & 1.000 & 17.128 & $1.0\times$ \\
\textit{Stop gradient} & blocked certificate gradient  & $0.0019 \pm 0.0004$ & 1.000 & 1.161 & $14.8\times$ \\
\textit{Sampled surrogate} & sampled grid loss  & $0.0024 \pm 0.0002$ & 1.000 & 16.223 & $1.1\times$ \\
\textit{Post hoc} & no certificate during training  & $0.0019 \pm 0.0004$ & 1.000 & 0.945 & $18.1\times$ \\
\textit{DSOS} & inner linear certificate  & 0.002 & 1.000 & 96.710 & $0.18\times$ \\
\textit{SDSOS} & inner second order certificate  & 0.002 & 1.000 & 114.952 & $0.15\times$ \\
\bottomrule
\end{tabular}%
}

\end{table}

\begin{table}[H]
\centering
\small
\caption{Coupled two pendulum chain schedule ablation.}
\label{tab:coupled_pendulum_schedule}
{%
\begin{tabular}{lrrrr}
\toprule
Schedule  & Cert.\ margin & Train time (s) & \makecell{Train speedup\\vs every step}  & Exact step frac. \\
\midrule
\textit{Every step} & $0.0036 \pm 0.0002$ & 17.128 & $1.0\times$ & 1.000 \\
\textit{Every 5} & $0.0025 \pm 0.0002$ & 1.490 & $11.5\times$ & 0.200 \\
\textit{Every 10} & $0.0025 \pm 0.0002$ & 1.361 & $12.6\times$ & 0.113 \\
\textit{Curriculum} & $0.0036 \pm 0.0002$ & 1.894 & $9.0\times$ & 1.000 \\
\bottomrule
\end{tabular}%
}

\end{table}

\begin{table}[H]
\centering
\small
\caption{Three pendulum short validation.}
\label{tab:three_pendulum}
{%
\begin{tabular}{lccccc}
\toprule
Mode & \makecell{Test cert.\\margin} & Cert.\ rate & \makecell{Exact\\step frac.} & Train time (s) & \makecell{Train speedup\\vs exact} \\
\midrule
\makecell[l]{\textit{Sparse SOS (Ours)}\\every step} & 0.0021 & 1.000 & 1.0 & 98.7 & $1.00\times$ \\
\textit{Curriculum} & 0.0021 & 1.000 & 1.0 & 97.9 & $1.01\times$ \\
\textit{Sampled surrogate} & 0.0014 & 1.000 & 0.0 & 97.4 & $1.01\times$ \\
\textit{Post hoc} & 0.0014 & 1.000 & 0.0 & 98.3 & $1.00\times$ \\
\textit{Stop gradient} & 0.0014 & 1.000 & 1.0 & 94.6 & $1.04\times$ \\
\bottomrule
\end{tabular}%
}

\end{table}

\subsection{Non-SOS structured precision verification}
\label{app:nonsos_full}

We test a related design principle outside polynomial SOS. It is not evidence for the sparse SOS optimization-layer backward, since the model uses a static PSD parameterization without differentiating through a conic solve. We include it only as breadth evidence for structured certificate parameterizations.

The task is Gaussian precision learning, where the model predicts an input-dependent precision matrix $K_\theta(x)$, the inverse covariance matrix of a Gaussian model. If $K_\theta(x)$ is not PSD, the Gaussian density is not well defined as a valid precision model. We use a synthetic eight-dimensional Gaussian whose true precision matrix has a chain sparsity pattern. The structured model predicts positive numbers $a_i$ for the diagonal terms and nonnegative numbers $w_i$ for edges between neighboring variables:
$K_\theta(x)
=
\operatorname{diag}(a_1,\ldots,a_8)
+
\sum_{i=1}^{7} w_i (e_i-e_{i+1})(e_i-e_{i+1})^\top ,
~ a_i>0,\; w_i\ge 0$ .
This parameterization is PSD by construction, since for any vector $y$,
$y^\top K_\theta(x)y
=
\sum_i a_i y_i^2
+
\sum_{i=1}^{7} w_i (y_i-y_{i+1})^2
\ge 0 $.
Thus the network outputs the sparse precision certificate variables $\{a_i\}_{i=1}^8$ and $\{w_i\}_{i=1}^7$ directly.

We compare against two standard alternatives. The \textit{post hoc repair} predicts a sparse chain matrix first and repairs it only after prediction when a valid matrix is needed. Since that repair is detached from the training computation, the model does not receive a gradient that says how to make its raw prediction certifiable. The \textit{dense Cholesky} baseline uses the standard PSD form $K=LL^\top$~\citep{golub2013matrix}. It is PSD in exact arithmetic since $y^\top LL^\top y=\|L^\top y\|^2$, but the dense factor $L$ does not preserve the sparse chain certificate measured by this verification.

\Cref{tab:nonsos} reports certificate quality only. The \textit{structured certificate} row certifies every test instance and has the largest sparse certificate margin. The \textit{dense Cholesky} row remains PSD as a dense matrix in exact arithmetic and a lower sparse certificate rate means that it does not expose the chain certificate family used by the structured model.

\Cref{tab:nonsos_full} shows the full results. Test negative log-likelihood (NLL) measures how well the predicted Gaussian distribution explains test observations and relative matrix error measures how accurately the model recovers the ground-truth precision matrix. Lower values indicate better data fit and more accurate matrix recovery, respectively. Sparse certificate margin and sparse certificate rate measure validity within the chain certificate family, where larger is better. The \textit{structured certificate} model has the strongest validity metrics. Its likelihood is competitive, but not uniformly best. This demonstrates that direct structured certificate prediction improves validity of the sparse PSD constraint.

\begin{table}[H]
\centering
\small
\caption{Structured precision check outside polynomial SOS.}
\label{tab:nonsos}
\begin{tabular}{lrr}
\toprule
Method & Sparse cert.\ margin & Sparse cert.\ rate \\
\midrule
\textit{Structured certificate (Ours)} & \textbf{0.5315} & \textbf{1.000} \\
\textit{Post hoc repair}        & 0.0373 & 0.705 \\
\textit{Dense Cholesky}         & 0.1674 & 0.905 \\
\bottomrule
\end{tabular}

\end{table}

\begin{table}[H]
\centering
\small
\caption{Structured precision learning outside polynomial SOS on five seeds.}
\label{tab:nonsos_full}
\begin{tabular}{lrrrr}
\toprule
Method & Test NLL & Rel.\ matrix error & Sparse cert.\ margin & Sparse cert.\ rate \\
\midrule
\textit{Structured certificate (Ours)} & 9.4311 & 0.3382 & \textbf{0.5315} & \textbf{1.000} \\
\textit{Post hoc repair}        & \textbf{9.3666} & 0.4860 & 0.0373 & 0.705 \\
\textit{Dense Cholesky}         & 9.4630 & 0.3893 & 0.1674 & 0.905 \\
\bottomrule
\end{tabular}

\end{table}

\subsection{Coupled certification-native strength and schedule}
\label{app:cert_native}

The certification-native checks compare sparse SOS with \textit{DSOS} and \textit{SDSOS} inner approximations on coupled control certificates. Here, \textit{certificate gradient} denotes training through the value of the selected certificate family, and \textit{sampled surrogate} denotes the sampled training objective. \Cref{tab:cert_strength} shows that all three certificate families can certify these small coupled instances, with order and cone choice changing margins and certificate time. \Cref{tab:cert_schedule_ablation} shows that exact certificate calls can be scheduled less frequently while retaining the same certification rate and nearly the same margin. These tables demonstrate that the control outcomes are not explained solely by one fixed schedule or one cone choice.

\begin{table}[H]
\centering
\small
\caption{Coupled certification-native strength comparisons.}
\label{tab:cert_strength}
{%
\begin{tabular}{llrrrrr}
\toprule
Training signal & Certificate family & Order & \makecell{Cert.\\margin} & \makecell{Cert.\\rate} & \makecell{Cert.\\time (s)} & \makecell{Cert. time\\rel.\ to sparse SOS} \\
\midrule
\textit{Certificate gradient} & \textit{DSOS} & 2 & 0.044 & 1.000 & 0.010 & $1.4\times$ \\
\textit{Certificate gradient} & \textit{DSOS} & 3 & 0.111 & 1.000 & 0.012 & $1.7\times$ \\
\textit{Certificate gradient} & \textit{SDSOS} & 2 & 0.052 & 1.000 & 0.008 & $1.1\times$ \\
\textit{Certificate gradient} & \textit{SDSOS} & 3 & 0.135 & 1.000 & 0.012 & $1.7\times$ \\
\textit{Certificate gradient} & \textit{Sparse SOS} & 2 & 0.052 & 1.000 & 0.007 & $1.0\times$ \\
\textit{Certificate gradient} & \textit{Sparse SOS} & 3 & 0.135 & 1.000 & 0.007 & $1.0\times$ \\
\textit{Sampled surrogate} & \textit{DSOS} & 2 & 0.014 & 1.000 & 0.006 & $1.2\times$ \\
\textit{Sampled surrogate} & \textit{DSOS} & 3 & 0.097 & 1.000 & 0.008 & $1.3\times$ \\
\textit{Sampled surrogate} & \textit{SDSOS} & 2 & 0.046 & 1.000 & 0.008 & $1.6\times$ \\
\textit{Sampled surrogate} & \textit{SDSOS} & 3 & 0.104 & 1.000 & 0.014 & $2.3\times$ \\
\textit{Sampled surrogate} & \textit{Sparse SOS} & 2 & 0.047 & 1.000 & 0.005 & $1.0\times$ \\
\textit{Sampled surrogate} & \textit{Sparse SOS} & 3 & 0.104 & 1.000 & 0.006 & $1.0\times$ \\
\bottomrule
\end{tabular}%
}

\end{table}

\begin{table}[H]
\centering
\small
\caption{Certification-native schedule ablation.}
\label{tab:cert_schedule_ablation}
{%
\begin{tabular}{lrrrrrr}
\toprule
Schedule & \makecell{Cert.\\rate} & \makecell{Cert.\\margin} & Train time (s) & \makecell{Train speedup\\vs every step} & \makecell{Cert.\\time (s)} & \makecell{Exact\\step frac.} \\
\midrule
\textit{every step} & 1.000 & 0.266 & 27.211 & $1.0\times$ & 0.007 & 1.000 \\
\textit{every 5} & 1.000 & 0.263 & 3.791 & $7.2\times$ & 0.007 & 0.203 \\
\textit{every 10} & 1.000 & 0.262 & 3.594 & $7.6\times$ & 0.008 & 0.103 \\
\textit{late-stage} & 1.000 & 0.262 & 4.079 & $6.7\times$ & 0.008 & $<0.001$ \\
\textit{curriculum} & 1.000 & 0.266 & 14.094 & $1.9\times$ & 0.007 & 1.000 \\
\bottomrule
\end{tabular}%
}

\end{table}

\subsection{Compute budget}
\label{app:compute_budget}

\Cref{tab:compute_budget_appendix} records the wall-clock budget on a single 20 thread Intel i9 9900X CPU for the learning studies. The table demonstrates that the diagnostics are modest in absolute compute, but the dense differentiable layers can still dominate runtime in the repeated optimization-layer experiments. 

\begin{table}[H]
\centering
\small
\caption{Compute budget summary.}
\label{tab:compute_budget_appendix}
\begin{tabular}{lrrrrr}
\toprule
Study & Seeds & Modes & Runs & Total budget (s) & Mean run (s) \\
\midrule
Coupled certification-native & 3 & 4 & 12 &  66.598 &  5.550 \\
Single pendulum control      & 5 & 3 & 15 & 461.084 & 30.739 \\
Two pendulum control         & 5 & 6 & 30 & 1235.595 & 41.187 \\
Three pendulum validation    & 1 & 5 & 5  & 486.900 & 97.380 \\
Non-SOS chain precision      & 5 & 3 & 15 &  59.510 &  3.967 \\
\bottomrule
\end{tabular}

\end{table}
\subsection{Framework implementation timings}
\label{app:stack_timings}
\Cref{tab:scaling_combinatorial,tab:scaling_lyapunov} give the small-size implementation measurements for banded binary optimization and oscillator certificates, the pose graph results appear in \Cref{tab:scaling_pose_graph}. They compare complete software stacks, their time differences combine representation, modeling, solver, and differentiation choices. The controlled comparisons in \Cref{tab:controlled} isolate those factors more directly.

\begin{table}[H]
\centering
\small
\caption{Banded binary optimization implementation times in seconds.}
\label{tab:scaling_combinatorial}
\begin{tabular}{rcccccc}
\toprule
$N$ & \multicolumn{3}{c}{Forward time (s)} & \multicolumn{3}{c}{Backward time (s)} \\
\cmidrule(lr){2-4}\cmidrule(lr){5-7}
& \makecell{\textit{Sparse SOS}\\\textit{(Ours)}} & \textit{CVXPY} & \textit{SDPR} & \makecell{\textit{Sparse SOS}\\\textit{(Ours)}} & \textit{CVXPY} & \textit{SDPR} \\
\midrule
$16$ & $0.018$ & $\mathbf{0.011}$ & $0.011$ & $\mathbf{3.80\!\cdot\!10^{-4}}$ & $5.24\!\cdot\!10^{-3}$ & $4.44\!\cdot\!10^{-3}$ \\
$32$ & $0.027$ & $0.071$ & $\mathbf{0.021}$ & $\mathbf{1.50\!\cdot\!10^{-3}}$ & $0.050$ & $0.022$ \\
$48$ & $0.037$ & $0.226$ & $\mathbf{0.036}$ & $\mathbf{3.38\!\cdot\!10^{-3}}$ & $0.188$ & $0.120$ \\
$64$ & $\mathbf{0.048}$ & $0.849$ & $0.048$ & $\mathbf{6.07\!\cdot\!10^{-3}}$ & $0.638$ & $0.434$ \\
$96$ & $\mathbf{0.072}$ & $5.307$ & $0.079$ & $\mathbf{0.014}$ & $3.583$ & $2.700$ \\
\bottomrule
\end{tabular}
\end{table}
\begin{table}[H]
\centering
\small
\caption{Coupled oscillator Lyapunov certificates implementation times in seconds.}
\label{tab:scaling_lyapunov}
\begin{tabular}{rcccc}
\toprule
$N$ & \multicolumn{2}{c}{Forward time (s)} & \multicolumn{2}{c}{Backward time (s)} \\
\cmidrule(lr){2-3}\cmidrule(lr){4-5}
& \makecell{\textit{Sparse SOS}\\\textit{(Ours)}} & \textit{CVXPY} & \makecell{\textit{Sparse SOS}\\\textit{(Ours)}} & \textit{CVXPY} \\
\midrule
$4$ & $\mathbf{1.66\!\cdot\!10^{-2}}$ & $9.38\!\cdot\!10^{-2}$ & $\mathbf{4.37\!\cdot\!10^{-4}}$ & $1.26\!\cdot\!10^{-1}$ \\
$6$ & $\mathbf{1.65\!\cdot\!10^{-2}}$ & $4.52\!\cdot\!10^{-1}$ & $\mathbf{9.10\!\cdot\!10^{-4}}$ & $9.12\!\cdot\!10^{-1}$ \\
$8$ & $\mathbf{1.86\!\cdot\!10^{-2}}$ & $1.87$ & $\mathbf{1.58\!\cdot\!10^{-3}}$ & $4.08$ \\
\bottomrule
\end{tabular}
\end{table}
\clearpage
\section{Scaling and Test Design Results}
\label{app:new_results}
\subsection{Large-scale numerical qualification}
\Cref{tab:large_accuracy} reports qualification at matched $\rho=0.45$, with all reported primal statuses optimal. At $N=256$, \textit{CVXPY} reports an optimal primal solve while diffcp reports \texttt{Solved/Inaccurate} for the subsequent differentiation solve. Tightening SCS does not resolve the dense discrepancy. \textit{SDPR} retains MOSEK defaults because its tested interface exposes no single control analogous to SCS's \texttt{eps}. Finite differences computed independently through two solvers agree from $1.47\times10^{-5}$ to $5.95\times10^{-4}$. We infer no cause for the tested derivative discrepancies.
\subsection{Clique width and formulation crossover}
\Cref{tab:width_full} gives the full sweep underlying \Cref{tab:width_crossover}. Reported ratios are ambient/chordal times. For random families, each cell summarizes ten independently generated instances. Its median ratio is the median across instances of ambient warm-call median divided by chordal warm-call median. Each banded cell summarizes ten timing repetitions of one deterministic instance. Quartiles and IQRs are reported at the precision of the experiment. All $200$ paired runs qualify. The width-eight ratio at $n=224$ has interquartile interval $[0.999,1.104]$, thus the crossover is an empirical median criterion for the tested families.

\begin{table}[H]
\centering
\small
\caption{Full width-controlled results.}
\label{tab:width_full}
\begin{tabular}{lrrrrrr}
\toprule
Family & $w$ & $n$ & Median & Q1 & Q3 & IQR\\
\midrule
Banded & 4 & 24 & 0.555 & 0.535 & 0.560 & 0.024\\
Banded & 4 & 48 & 0.635 & 0.633 & 0.636 & 0.004\\
Banded & 4 & 96 & 1.068 & 1.063 & 1.070 & 0.007\\
Banded & 4 & 160 & 1.807 & 1.804 & 1.812 & 0.008\\
Banded & 4 & 224 & 3.084 & 3.080 & 3.088 & 0.008\\
\midrule
Random chordal & 4 & 24 & 0.467 & 0.451 & 0.495 & 0.044\\
Random chordal & 4 & 48 & 0.626 & 0.600 & 0.659 & 0.060\\
Random chordal & 4 & 96 & 1.076 & 1.038 & 1.139 & 0.101\\
Random chordal & 4 & 160 & 1.894 & 1.772 & 1.965 & 0.193\\
Random chordal & 4 & 224 & 3.071 & 2.975 & 3.184 & 0.210\\
\midrule
Random chordal & 6 & 24 & 0.288 & 0.280 & 0.294 & 0.015\\
Random chordal & 6 & 48 & 0.363 & 0.351 & 0.367 & 0.016\\
Random chordal & 6 & 96 & 0.614 & 0.613 & 0.621 & 0.008\\
Random chordal & 6 & 160 & 1.057 & 1.038 & 1.070 & 0.032\\
Random chordal & 6 & 224 & 1.734 & 1.696 & 1.758 & 0.062\\
\midrule
Random chordal & 8 & 24 & 0.209 & 0.206 & 0.212 & 0.007\\
Random chordal & 8 & 48 & 0.255 & 0.252 & 0.257 & 0.005\\
Random chordal & 8 & 96 & 0.392 & 0.383 & 0.402 & 0.019\\
Random chordal & 8 & 160 & 0.667 & 0.645 & 0.706 & 0.062\\
Random chordal & 8 & 224 & 1.047 & 0.999 & 1.104 & 0.105\\
\bottomrule
\end{tabular}
\end{table}
\clearpage
\subsection{Per-seed design outcomes}
\label{app:design_outcomes}
The experiment in \Cref{tab:design} uses $16/6/4/6$ instances per seed for training, test instances, $n=96$, and $n=160$, respectively. The same test instances are reconstructed for each seed. \Cref{tab:design_seeds} reports the individual seed counts, with \textit{stop gradient} and \textit{no layer} matching these denominators for every seed. \Cref{tab:design_margins} reports the continuous certified margins against the requirement $\delta=0.005$, with SD computed as the sample standard deviation of the five seed means and the worst margin taken over all evaluated designs. At $n=160$, failures of \textit{finite differences} and the \textit{sampled surrogate} are concentrated in complete seed cells.

\textit{Sparse SOS}'s two smallest seed-mean slacks are $0.0013$ and $0.0014$, and its worst design slack is $0.0006$. The \textit{sampled surrogate} has the largest mean margin but also the largest variability and a negative worst margin. \textit{Sparse SOS} is the only method with zero violations on every split and seed, while the \textit{sampled surrogate} meets the requirement on three of the five seeds.

\begin{table}[H]
\centering
\small
\caption{Violation counts for all seeds. }
\label{tab:design_seeds}
\begin{tabular}{lccc}
\toprule
Split & \makecell{\textit{Sparse SOS}\\\textit{(Ours)}} & \textit{Finite differences} & \textit{Sampled surrogate}\\
\midrule
Train & 0, 0, 0, 0, 0 & 0, 0, 3, 0, 1 & 0, 0, 0, 0, 9\\
Test instances & 0, 0, 0, 0, 0 & 0, 0, 0, 0, 0 & 0, 0, 0, 0, 4\\
$n=96$ & 0, 0, 0, 0, 0 & 0, 4, 0, 0, 0 & 0, 0, 1, 0, 4\\
$n=160$ & 0, 0, 0, 0, 0 & 0, 6, 0, 0, 0 & 0, 0, 6, 0, 6\\
\bottomrule
\end{tabular}
\end{table}
\begin{table}[H]
\centering
\small
\caption{Continuous certified relative margins at $n=160$. }
\label{tab:design_margins}
\begin{tabular}{lrrr}
\toprule
Training signal & Mean of seed means & SD & Worst design\\
\midrule
\textit{Sparse SOS (Ours)} & 0.0114 & 0.0051 & 0.0056\\
\textit{Finite differences} & 0.0113 & 0.0184 & $-0.0146$\\
\textit{Sampled surrogate} & 0.0205 & 0.0333 & $-0.0062$\\
\textit{Stop gradient} or \textit{no layer} & $-0.0059$ & 0.0018 & $-0.0082$\\
\bottomrule
\end{tabular}
\end{table}

\clearpage
\end{document}